# Keypoint Description by Symmetry Assessment–Applications in Biometrics

Anna Mikaelyan, Fernando Alonso-Fernandez, and Josef Bigun, *Fellow, IEEE*

*Abstract*—We present a model-based feature extractor to describe neighborhoods around keypoints by finite expansion, estimating the spatially varying orientation by harmonic functions. The iso-curves of such functions are highly symmetric w.r.t. the origin (a keypoint) and the estimated parameters have well defined geometric interpretations. The origin is also a unique singularity of all harmonic functions, helping to determine the location of a keypoint precisely, whereas the functions describe the object shape of the neighborhood. This is novel and complementary to traditional texture features which describe texture-shape properties i.e. they are purposively invariant to translation (within a texture). We report on experiments of verification and identification of keypoints in forensic fingerprints by using publicly available data (NIST SD27), and discuss the results in comparison to other studies. These support our conclusions that the novel features can equip single cores or single minutia with a significant verification power at 19% EER, and an identification power of 24-78% for ranks of 1-20. Additionally, we report verification results of periocular biometrics using near-infrared images, reaching an EER performance of 13%, which is comparable to the state of the art. More importantly, fusion of two systems, our and texture features (Gabor), result in a measurable performance improvement. We report reduction of the EER to 9%, supporting the view that the novel features capture relevant visual information, which traditional texture features do not.

## I. INTRODUCTION

WE suggest the use of a model explaining the orientation field of the neighborhood of a keypoint by a finite sequence of basis functions which have a singularity in common, the keypoint itself. The purpose is to give the keypoint an identity as much unique as possible by explaining its neighborhood with reference to the singularity. This is because each basis function (except one) has no other singularity than the one at the origin, which is the keypoint, easing identification of keypoints.

Because it is model-based, the feature extraction process has the power of providing information about the quality of the model fit. Model parameters are the features that explain a neighborhood. The explanation can be significant or poor, which is represented by the quality.

The goal of the description is to represent the properties of the neighborhood as unique as possible, to give to its center a label, the keypoint identity. Furthermore, we want to give an identity to few points in comparison to texture segmentation where one wishes to give an identity or label to a continuum of points, a region. Thus the goal is to extract *object properties* rather than *texture properties* of a neighborhood[1].

We will study our descriptors in the context of forensic fingerprints (500 dpi), Fig. 1. They are comprised of ridges forming singularities in micro scale (minutiae, which can be of *type* bifurcation or end-point), or macro-scale (cores, and deltas) [1]. Collectively, we will call them *keypoints*. The fingerprints taken at highly controlled environments such as police-stations are called *tenprints*. Fingerprints originated from uncontrolled conditions e.g. collected from a crime scene are called *fingermarks* (in Europe), or *latents* (in USA). Fingermarks have very poor quality compared to tenprints, posing challenges to human and machine experts alike.

We have also studied our feature extraction by verifying identity using periocular images. Irises are usually not distinguished from one another by extracting keypoints in biometric recognition, but by methods quantifying texture properties. We evaluated if object properties bring complementary information to periocular recognition by extracting them on a regular grid of points placed at the pupil center.

### A. Related work

A desired property of texture features is invariance to translation by which the pixels of a region inherit a common property allowing to delineate the texture from other textures, [2]. However, machine vision also uses sparse *keypoints* to which the corresponding feature vectors are associated. In combination, such keypoints can be used to identify *visual objects*, for e.g. image content based retrievals, [3], navigation [4], and image registration [5]. The feature vector describes then the neighborhood around the keypoint it is associated with. With this in mind, several feature vectors have been suggested [6], including Scale-invariant feature transform (SIFT) [7] or Speeded Up Robust Features (SURF) [8] for general visual object recognition.

One of the earliest usages of image comparison by keypoints is in forensic fingerprint matching, long before the computer era, e.g. the 19'th century contributors, J. Purkyně, W. Herschel, A. Bertillon, F. Galton, E. Henry, A. Haque, C. Bose, e.g. [9]. Here the object is a finger and the mission is to conclude if two fingerprints originate from it, by using keypoints which are minutiae, cores, deltas. General purpose

---

[1]This is an analogy of "particle" and "wave" notions in physics, where the former is characterized by a well defined position, and the latter by being repetitive.

The authors are with Halmstad University, IDE dept, SE-30234, Sweden. E-mail: see http://islab.hh.se/mediawiki/Personnel

keypoint descriptors such as SIFT [10], SURF [11], LBP [12] have hitherto not been as performant as fingerprint specific descriptors. Nonetheless, in matching tenprints to tenprints, SIFT features are suggested to extract texture properties, reaching ∼10% EER, in contrast to minutiae positions and directions descriptor performance of 2% EER [10].

Similar to SIFT, LBP applies histogramming, which is the source of their translation invariance, to binary codes representing the orientations of iso-curves passing through a keypoint and a circle around it. The latter is, in its essence, what makes SIFT and SURF features translation invariant (texture descriptors) too. Performance of the texture based periocular and iris images with these features is therefore expectedly high: 7% EER with SIFT features, 19% with LBP [13] and ∼11% EER with SIFT features [14].

To the best of our knowledge there is no study on the performance of general purpose feature vectors when matching fingermarks to tenprints. This is presumably because, i) generic feature vectors are most efficient when applied to *own* keypoints (rather than minutiae), ii) they extract 2-3 orders of magnitudes more keypoints than what a human fingerprint expert endorses as reliable, iii) the repeatability of their extracted keypoints on fingermarks are yet to be demonstrated and iv) human expert cannot interpret or interfere with the high (128 in [10], [11]) dimensional vectors for each keypoint.

For good quality fingerprints Gabor filter responses at (8) different directions can be established [15], yielding the fingercode of the neighborhoods of a core. The study of [16] is similar in the spirit and suggests a polar sampling of the gradient field (angle, varying in $[0, \pi]$) around a keypoint to be neighborhood descriptors. By contrast works [17], [18] reports *identification* performance of fingermarks against tenprints using minutiae directions, skeleton, and orientation fields, showing that there is a significant unexploited potential of non-minutia features in identification. However, the orientation fields and (ridge) skeletons of fingermarks are reconstructed from manually extracted minutiae, whereas those of the tenprints were based on outputs of (undisclosed) commercial software.

Periocular recognition has gained attention recently in the biometrics field [19], [20], [13], [21]. Periocular refers to the face region in the immediate vicinity of the eye, including the eye, eyelids, lashes and eyebrows. It has emerged as a promising trait for unconstrained biometrics, with a surprisingly high discrimination ability. One advantage is its availability over a wide range of distances even when the iris texture cannot be reliably obtained (low resolution) or under partial face occlusion (close distances). The most widely used approaches for periocular recognition include LBP and, to a lesser extent, Histogram of Oriented Gradients (HOG) [22] and SIFT keypoints. The use of different experimental setups and databases make difficult a direct comparison between existing works. The study of Park *et al.* [13] compares LBP, HOG and SIFT using the same data, with SIFT giving the best performance 6.95% EER, followed by LBP 19.26% EER and HOG 21.78% EER. Other works with LBPs, however, report EER below 1% [23], [24]. Gabor features were also proposed in a work of 2002 [19]. Here, the authors used three machine experts to process Gabor features from the facial regions surrounding the eyes and the mouth, achieving very low error rates(EER≤0.3%). Another important set of research works have concentrated their efforts in the fusion of different algorithms, for example [25], [26].

*B. Overview and contributions*

The feature extraction method includes three main steps as shown on Fig. 1. Given an input image, we estimate its orientation field, Sec. II. This is an iterative process for noisy images, which also includes the absolute frequency field as a byproduct. The feature extractor expects that *its input* has complex (pixel) values, where argument (angle) and magnitude (real, non-negative) information define the model parameter and the error of fitting. This is in itself not novel, [27], [28], but automatic extraction of fingermark orientation fields (including the quality measures) and offering this field to a forensic (human) expert to verify or edit it, to the best of our knowledge, is a novelty. At the end of Step 1 (and even Step 2) the resulting complex fields are meaningful (for a human forensic expert) because the complex pixels are measurements of angles (model parameters) which can be displayed as a color image in HSV color space by steering the hue component (orientation angle) and the brightness component (quality), respectively. Usage of the same complex representation offers *higher resolution* both for the human, seeing a color image, and an algorithm, handling a dense complex field. The human examiner can interpret dark pixels as low-confidence pixels and clearly visible hue as reliable angle parameters. This enables an interface between the forensic examiner, and a machine algorithm, for manual verification or editing of the computed angle estimations.

At Step 2 we confine the dense complex field around an arbitrary keypoint to a set of torus shaped areas of growing radii. This is done by multiplying the complex valued image by the (non-negative, real) magnitudes of a set of filters which are torus shaped and are normalized to reflect the quality as well as absence of quality within the support areas of filter functions, Sec. II. Subsequently, the extracted complex tori are projected onto a set of complex filter functions, which fit angle parameters of highly symmetric function families to the input (complex) tori, along with the quality of the fitting, Sec. III-E. The result is equivalent to extracting a Generalized Structure Tensor feature by using the (complex) symmetry derivatives of Gaussians. Neither of the latter concepts are novel, [29]. However, the filter function, Sec. III-C, realizing a (mathematically) dense set of filters, is a useful novelty. This is because, beside allowing finite expansion and extracting low-dimensional features which are meaningful to humans, the filters can be easily adapted to novel applications, via their parameters, Sec. III-B. The latter are directly connected to width and location of tori, as detailed in the Appendix.

The resulting feature vector explains the image content around a keypoint by families of functions which all except one have the keypoint as singularity. This makes the feature vector as a descriptor of object rather than texture properties of the keypoint neighborhood, because the filter response

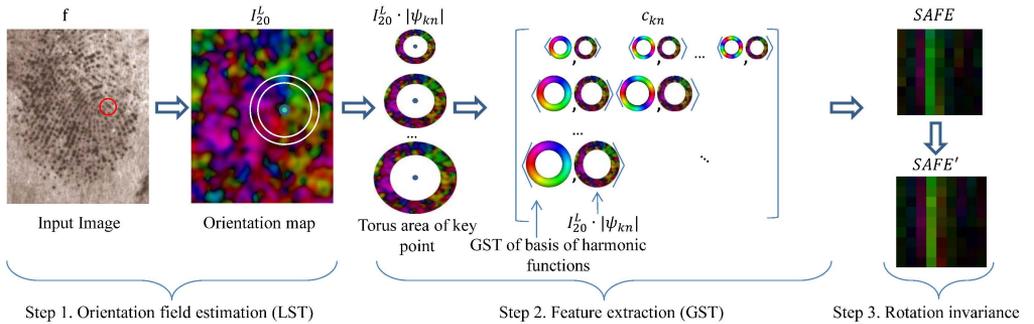

Fig. 1. Flow chart of the suggested object-based feature extraction visualized by the example of fingerprint image. *Step 1*. Preprocessing by means of Linear Structure Tensor (LST) to produce dense orientation field. *Step 2*. Extraction of ring shaped areas in neighbourhood of point of interest followed by feature extraction. Extraction is performed by projecting orientation image information on basis of harmonic functions generated with Generalized Structure Tensor (GST). *Step 3*. Alignment by rotation compensation in the feature space for matching

magnitudes quickly attenuate when going away from the keypoint. We are not aware of other feature vectors that systematically measure image contents that are not translation invariant. The proposed features are *translation variant*, and rotation steerable, achieving *rotation invariance* by rotation compensation, through complex multiplication, Step 3, Sec. III-D. This is important to practice since rotation invariance is achieved by few multiplications of the features, without rotating the underlying image data. The steerable filter theory of [30], which is linear w.r.t. the original image, shares some concepts with the feature vector presented in Sec. III-C. Nonetheless it is significantly different, since the steerability of our descriptors concerns orientation field, which is *non-linear*, though tractably[2] connected to the iso-curves of the original image, [29]. The arguments of our feature vector elements are angle parameters of harmonic functions and the magnitudes are quality measures, which both can be interpreted visually by humans as collection of curves. We are not aware of other feature vectors which are orientation steerable and represent visually meaningful curves.

In the experiments, Sec. IV, we provide verification and identification results using publicly available databases, and published methods, for repeatability and future comparisons, to quantify the recognition power of the suggested features in isolation. This is novel in its essence for fingerprints because in prior studies it is not possible to read out the recognition power of descriptors for a variety of reasons. These include i) reporting only identification performance (CMC-curves) means that the background data must be fully available and future experiments must actually precisely use them[3], ii) all used methods are not published in prior studies, which is an obstacle to a critical analysis, e.g. on their influence in the suggested recognition performance in comparison to that of the descriptors.

[2]Squaring is implicit in the structure tensors or their equivalent complex fields representing orientations in double angle.

[3]The CMC-curves are "normalized" with the number of people in the query population, not with the number of people in the background data. Accordingly, the curves will systematically shift with the size of the background data.

## II. ORIENTATION FIELD ESTIMATION

Our feature extractor analyzes variations of the complex valued orientation field around keypoints. We suggest to use the Linear Symmetry Tensor (LST) [27] with the purpose to obtain a dense orientation field together with quality measures. However, to obtain a reliable orientation field, an iterative procedure can be utilized if the input image is noisy, as is the case with forensic fingermarks. The procedure comprises of an initial estimation of the dense orientation field and the (absolute) frequency field improving one another in subsequent iterations via enhanced images in intermediate steps [28].

The ordinary structure tensor (ST) [27] is fully equivalent to one complex, denoted $I_{20}$, and one real scalar $I_{11}$, which in sequel define a vector, LST, the Linear Symmetry Tensor

$$LST = \begin{pmatrix} I_{20} \\ I_{11} \end{pmatrix} = \begin{pmatrix} (\lambda_{max} - \lambda_{min})e^{2i\angle k_{max}} \\ \lambda_{max} + \lambda_{min} \end{pmatrix} \quad (1)$$

where $\lambda_{\min}, \lambda_{\max}$ are the minor and major eigenvalues of the tensor. Equation (1) asserts that the predominant direction of the neighbourhood $f$, represented by the major eigenvector $k_{max}$ of ST, is directly encoded in the argument of $I_{20}$, albeit in double angle representation of the former. Contrast, or edge energy of the neighborhood is obtained by $I_{11}$. Written in polar form, the complex component of $I_{20}$ can be displayed as a color/hue in HSV colour space modulating (twice) the direction angle of dominant orientation, whereas brightness of pixels modulate the magnitude, Fig. 2 *Middle*.

The degree of consistency between gradient vectors involved in the orientation estimation is read in $|I_{20}|$, which in turn can be at most $I_{11}$ for perfect orientation fit, allowing for normalization

$$I_{20}^L = I_{20}^{(0)}/I_{11}^{(0)}, \text{ where } |I_{20}^L| \leq 1. \quad (2)$$

It is known that $I_{20}$ and $I_{11}$ can directly be obtained by averaging (Gaussian) squares of complex gradients, [27], as summarized here next. When modeling and measuring other symmetries than linear in image neighborhoods, a similar method can be applied but using complex filters (instead of Gaussian), [29], having integer indexes describing their phase component. The superscript 0 referring to the linear symmetries (described by the ordinary ST or LST) is thus used to avoid confusion with descriptors of other symmetric patterns, presented in sequel.

The dense orientation image $I_{20}^L$ can computationally be obtained in three steps. First the original image is convolved with a member (n = 1) of a filter family called the symmetry derivatives of Gaussians, [5], and then squared (pixel wise). The filter family is defined as follows

$$\Gamma^{\{n,\sigma^2\}} = (D_x + iD_y)^n e^{-\frac{r^2}{2\sigma^2}} = r^n \frac{1}{\kappa_n} e^{-\frac{r^2}{2\sigma^2}} e^{in\varphi}. \quad (3)$$

Here the constant $\kappa_n$ assures that the norm of the filter is 1, whereas $r = |x + iy|$ and $\varphi = \angle(x + iy)$.

The result is a complex image and is called Infinitesimal Linear Symmetry Tensor (ILST)

$$ILST = (\tilde{I}_{20}^{(0)}, |\tilde{I}_{20}^{(0)}|)^T, \text{ with } \tilde{I}_{20}^{(0)} = (\Gamma^{\{1,\sigma_{in}^2\}} * f)^2 \quad (4)$$

This definition allows to formulate the second step of LST as a linear filtering of ILST

$$LST = \begin{pmatrix} \Gamma^{\{0,\sigma_{out}^2\}} * \tilde{I}_{20}^{(0)} \\ |\Gamma^{\{0,\sigma_{out}^2\}}| * |\tilde{I}_{20}^{(0)}| \end{pmatrix} = \Gamma^{\{0,\sigma_{out}^2\}} * ILST \quad (5)$$

It is worth noting that two different scale parameters are involved: inner scale $\sigma_{in}^2$ depending on the spatial frequency content of the neighborhood and the outer scale $\sigma_{out}^2$ which defines the size of the neighborhood.

Many images, e.g. fingermarks, are notoriously noisy and the orientation fields are difficult to obtain automatically. One of the reasons is that the local (absolute) frequency corresponding to parameter $\sigma_{in}^2$ varies with image location. It has been shown that if the inner scale $\sigma_{in}^2$ is changed in discrete (but not necessarily uniform) steps, the orientation of $\log(I_{11}^{(0)})$ can be invertibly mapped to the frequency, [28]. Accordingly, even dense frequency fields can be obtained by applying the LST, but to the logarithmic scale space of the contrast $I_{11}^{(0)}$ since LST is an orientation fitting tensor.

The corresponding LST for estimating the frequency map from the sampled logarithmic scale space consists of a complex valued $I_{20}$ and a real valued $I_{11}$, with normalized orientation of frequency map equaling to $I_{20}^F = I_{20}/I_{11}$. Thus, the signal representation of the frequency $I_{20}^F$ is identical to that of the orientation $I_{20}^L$, except that its argument encodes the absolute frequency of the neighborhood. We obtained dense orientation and (absolute) frequency fields for fingerprint images, Fig. 2, through an iteration process [28]. The originals are a genuine tenprint-fingermark pair, left column. The orientation fields are illustrated by the middle images. The frequency fields are shown by the right column with frequency range varying in accordance with how the color progresses in rainbow. For "mnemonic" reasons the straightforward frequency visualization was updated such that ridge periods correspond to increasing electromagnetic wavelengths. For human perception it is more natural to identify increasing range with violet–red colour palette without the need to remember that, for example, red corresponds to lower frequency as compared to blue/violet. The frequency field contains less hue variations because not all possible frequencies are present with sufficient prominence, in contrast to orientation field.

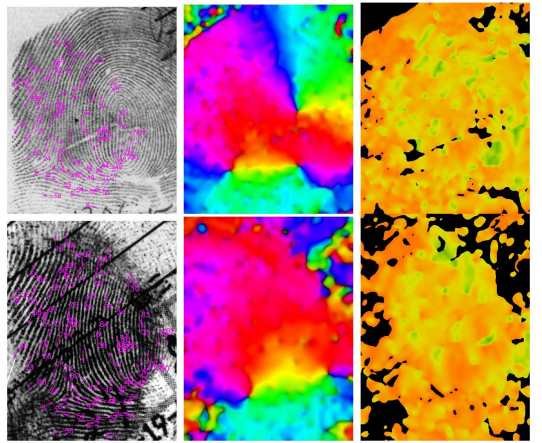

Fig. 2. Orientation and frequency dense maps of matched pair of good quality tenprints and low quality noisy forensic fingerprint. Hue represents orientation angle and value/intensity represents the certainty of measured angle in HSV colour space.

## III. ORIENTATION FIELD DESCRIPTORS

The Generalized Structure Tensor (GST) is an extension of the LST for more elaborate symmetries, especially for assessing orientations defined by iso-curves of harmonic functions. By using an integer $n$ it is possible to obtain a taxonomy of the symmetry types (linear, parabolic, spiral, hyperbolic, etc.) and their associated orientations, with $n = 0$ corresponding to LST. GST is defined in a similar way as LST

$$GST(n) = (I_{20}^{(n)}, I_{11}^{(n)})^T = \Gamma^{\{n,\sigma_{out}^2\}} * ILST \quad (6)$$

and fits an iso-curve chosen from the function family fixed by $n$ [31]. In Fig. 3 *First Row* one pattern of each family is shown by changing $n$. For $n \neq 0$ the filter function $\Gamma^{\{n,\sigma_{out}^2\}}$ introduces a complex filter[4], whose magnitude is a circular torus (ring), and argument is an integer power of $\varphi$. In Fig. 3 *Third Row* brightness represents the filter magnitude and hue its argument whereby the frequency of the same hue is determined by $n$.

However, the $\sigma_{out}$ we suggest in GST are considerably larger than the ones used in LST to obtain the orientation and frequency fields, Sec. II. The $\sigma_{out}$ used in LST is small enough to ensure local linearity of the iso-curves, i.e. every complex pixel in the orientation map of Fig. 1 is based on a region with the same size as the red circle shown in the original. This is to be compared to the size of $\sigma_{out}$ used in GST, the region between the white circles of the orientation map, whose goal is to capture sufficient orientation or frequency variation around the key point, (green), to give a unique "character" to it.

In GST, the symmetry derivative filters define an orthogonal spanning set for orientation fields in the angular direction $\varphi$,

---
[4]Note that in LST, where $n = 0$, this filter is real, being a Gaussian

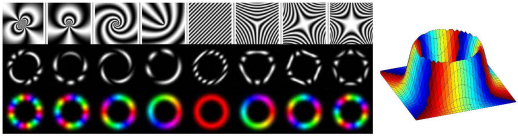

Fig. 3. *Top:* Sample patterns of the family of harmonic functions used as basis. Patterns are displayed for $\theta = \pi/4$, [32]. *Second:* One pattern per original (top), but in selected tori support $|\psi_{kn}|$. *Third:* Filters used to detect patterns above, with $n = -4:3$. *Right:* A sample filter $\psi_{k,-4}$ shown in 3D, with color representing $\angle \psi_{kn}$ whereby every hue appears 4 times.

eq. (3), Fig. 3 *Second Row* (but not radially yet). They will be used to define a complete set of basis functions on which the orientation image $I_{20}^L$ will be projected (Sec. III-A-III-C). The coefficients will then explain the angular variation of the orientation field in rings around the keypoint.

Each filter $\Gamma^{\{n,\sigma_{out}^2\}}$ matches to the orientation field of a member of the family defined by $n$, Fig. 3 *First row*. A family is characterized in that its members must have iso-curves which can be described by a distinct parameter $n$, which represents the direction of lines in harmonic functions. For example, the iso-curves of the spirals in Fig. 3 *First Row* are due to lines in *log-polar space*, $\cos(\theta)\Re(\log z) + \sin(\theta)\Im(\log z) = const$, with $z = x + iy$ and $\theta = \pi/4$ [5]. The iso-curves of the other patterns have similar explanations by way of precise *curvilinear coordinates* of harmonic functions which are detailed elsewhere [29]. As in LST, the orientation parameter $\theta$ of GST is estimated and represented in double angle via $2\theta = \angle I_{20}^{(n)}$, to achieve uniqueness. Only one member per family is shown in the figure, and all members are deliberately chosen to have the same parameter $2\theta = \pi/2$ albeit in different families denoted by $n = -4 \cdots 3$.

The GST delivers thus $2\theta$ in the argument of $I_{20}^{(n)}$ and an evidence of the presence of the symmetry follows from the ratio $|I_{20}^{(n)}|/I_{11}^{(n)} \leq 1$, which is brightness independent. Similarly to linear symmetry, the inequality relationship holds as equality iff the fitting of $\theta$ is error free. If we vary the parameter $\theta$ with the increment $\Delta\theta$, all image patterns at the top-row would rotate proportionally, with the exception of the spiral pattern. The latter, which corresponds to $n = -2$, would first become tighter, then turn into circles and finally will change the sense of twist, when its $\theta$ changes.

The support of the filter $|\Gamma^{\{n,\sigma_{out}^2\}}|$ defines the support of the region in which the orientation is modelled. Spatial support is independent of the symmetry index $n$, offering an opportunity for completeness radially (in addition to the angular coordinate) by changing the support of the filter. Thus increase of the number of basis functions will be enabled in two ways, angularly and radially. This will allow description of general complex fields, e.g. contrast normalized orientation fields $I_{20}^{(0)}/I_{11}^{(0)}$ and unnormalized orientation fields $I_{20}^{(0)}$ densely by finite expansion up to a prescribed error.

---

[5] The symbols $\Re$ and $\Im$ are real and imaginary parts respectively.

## A. Finite Expansion of Orientation Fields Angularly

Choosing the keypoint as origin, we suggest usage of a sequence of GSTs, Fig. 3 *Third Row*, to describe the orientation field $h(x, y)$ around the keypoint, Fig. 1 *Step 2*

$$h(x,y) = \sum_k h_k(x,y) \quad \text{with} \quad h_k(x,y) = \sum_n c_{kn}\psi_{kn} \quad (7)$$

where

$$\psi_{kn} = \frac{1}{\kappa_k} r^\mu e^{-\frac{r^2}{2\sigma_k^2}} e^{-in\varphi}. \quad (8)$$

The normalization constants $\kappa_k$ are chosen to assure $\|\psi_{kn}\| = 1$. Note that $|\psi_{kn}|$ is independent of $n$ due to magnitude $|\cdot|$ operation although $\psi_{kn}$ does depend on $n$. Additionally, if consequent tori areas $h_k$ and $h_{k+1}$ have negligible overlap, then filters $|\psi_{kn}|$ will be nearly orthogonal. These tori are steered by $\mu$ and $\sigma_k$ as detailed in the next section and the Appendix.

For every $h_k$, which is the orientation field confined to a torus, it is possible to associate a finite array of complex coefficients $c_{kn}$ by varying the symmetry index of the filter $n$

$$c_{kn} = <\psi_{kn}, h_k> = \int \psi_{kn}^* h_k \quad (9)$$

where $<,>$ denotes the ordinary scalar product defined as the shown (double) integral. Coefficients $c_{kn}$ are analogous to $I_{20}^{(n)}$, except that the radial support of the integral, $|\psi_{kn}|$, is controlled more precisely (by $k$). In return the coefficients can be used to synthesize $h_k = \sum_n c_{kn} e^{-in\varphi}$ up to a prescribed error ($\mathcal{L}^2$), which follows from the orthogonality of $\psi_{kn}$ and Fourier theory. In addition we make sure that $h_k$ is not constant in the radial direction by letting the thickness of $\psi_{kn}$ be as small as required by adapting its internal parameters. The coefficients $c_{kn}$ will then tell the amount of the corresponding angular basis $e^{-in\varphi}$ that is present in the orientation field $h_k$

Analogously, the coefficients

$$e_k = <|\psi_{kn}|, |\psi_{kn}| \cdot |I_{20}^L|> \quad (10)$$

correspond to $I_{11}^{(n)}$. As in the case of $c_{kn}$, the measurement delivered by this integral originates from the support of $|\psi_{kn}|$ in the orientation map, which is steered via $k$.

By way of example, we have estimated the projection coefficients $c_{kn}$ of the orientation field in Fig. 4 *Right*, via eq. (9). The orientation map is a field generated by

$$h(r,\varphi) = w_1^2 e^{-i\frac{\pi}{2}} r^{-1} e^{i\varphi} + 2w_1 w_2 + w_2^2 e^{i\frac{\pi}{2}} r e^{-i\varphi} \quad (11)$$

where the origin is the center of the image which is $257 \times 257$. The coefficients $w_1, w_2$ are two real constants. The orientation field is our input, and the equation is thus the ground-truth at every possible "thin" (concentric) torus. The orientation field is a linear combination of those of the component iso-curves, $n = -1$ and $n = 1$, and an additive constant field, $2w_1w_2$.

It is possible to invert the orientation field analytically to obtain the original[6], yielding

$$f(r,\varphi) = \cos(\Re[w_1 e^{-i\frac{\pi}{4}}(2r^{\frac{1}{2}} e^{i\frac{\varphi}{2}}) + w_2 e^{i\frac{\pi}{4}}(\frac{2}{3}r^{\frac{3}{2}} e^{i\frac{3\varphi}{2}})]). \quad (12)$$

---

[6] Obtaining $f$ from $h$ (inversion) in $((D_x + iD_y)f)^2 = h$ can be done by using the relationship between gradients of harmonic functions and complex derivation, *Appendix* of [29].

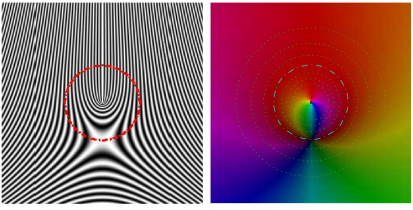

Fig. 4. *Left:* The original image as linear combination of parabolic and hyperbolic patterns *Right:* Orientation map of the original described with finite expansion coefficients.

| | $n=-1$ | $n=0$ | $n=1$ | $\epsilon_{-1}$ | $\epsilon_0$ | $\epsilon_1$ |
|---|---|---|---|---|---|---|
| | (1.23 4.71) | (1.23 6.28) | (0.31 1.57) | 0.002 | 0.000 | 0.001 |
| | (0.98 4.71) | (1.23 6.28) | (0.39 1.57) | 0.002 | 0.000 | 0.002 |
| | (0.78 4.71) | (1.23 6.28) | (0.49 1.57) | 0.001 | 0.000 | 0.002 |
| | (0.62 4.71) | (1.23 6.28) | (0.62 1.57) | 0.001 | 0.000 | 0.003 |
| | (0.49 4.71) | (1.23 6.28) | (0.78 1.57) | 0.001 | 0.000 | 0.004 |
| | (0.39 4.71) | (1.23 6.28) | (0.98 1.57) | 0.001 | 0.000 | 0.005 |
| | (0.31 4.71) | (1.23 6.28) | (1.24 1.57) | 0.000 | 0.000 | 0.006 |

Fig. 5. *Left:* Complex coefficients $c_{kn}$, where rows are for rings with increased radii, $k$ and columns for different symmetry. *Right:* The same but numerically in polar format $|c_{kn}|$ for the 3 mid-columns, and the corresponding error magnitudes. Groundtruth phases are $-\pi/2, 2\pi$ and $\pi/2$.

which is illustrated in Fig. 4 *Left*, whereby the link between line orientations and colors is immediate. The iso-curves of the original consist of a linear combination of iso-curves of the members of pattern families, resembling cores, $n = -1$ (or $z^{1/2}$ in polar coordinates, [29]), and deltas, $n = 1$ ($z^{3/2}$) of fingerprint images, Fig. 3.

From the equation of the orientation field and the corresponding gray scale image it can be seen that outside of a torus, indicated in the original, $|z| < r_0$ (where $r_0 = 64$), orientations of a delta like pattern dominate, whereas inside, the orientations are best described by those of a core pattern. On the torus itself the orientation schemes originating from the core and the delta have the same strength. Purposively $w_1$ and $w_2$ are chosen such that they generate the same (absolute) spatial frequencies on the indicated ring (*Red*, or *Dash-dot*), i.e. $w_1 r_0^{-1/2} = w_2 r_0^{1/2} = 2\pi/8 = \omega_0$ which is a typical frequency for fingerprints. This results in that the ground-truth for non-zero coefficients $[c_{-1}, c_0, c_1]^T$ are given by $[\mathbf{1} \cdot w_0^2 e^{-i\pi/2}|z|^{-1}, \mathbf{2} \cdot \omega_0^2, \mathbf{1} \cdot \omega_0^2 e^{-i\pi/2}|z|]$ on any (concentric) torus of the orientation field. It is thus the function

$$h_k(z) = |\psi_k.|h(z) = \frac{1}{\kappa_k}|z|^\mu e^{-\frac{r^2}{2\sigma_k^2}} h(z) \quad (13)$$

which will be described by the projections of (9).

The experimentally estimated coefficients are given as columns of table of Fig. 5. The magnitude values $|c_{kn}|$ are shown for 7 tori placed at exponentially increasing distances from the center, *Right*, as rows of table. The left half of the table shows the approximations of $c_{kn}$ *in polar coordinates* corresponding to the 3 "mid" basis functions, $._n$ with $n = -1, 0, 1$ only, since the rest were negligibly close to 0, as they should. This is illustrated by the $7 \times 8$ (estimated) coefficient matrix shown as a color image, where the coefficient magnitudes and argument angles modulate the brightness, and the hue respectively. The right half of the table shows the absolute error compared to the ground truth which confirm possibility to estimate coefficients accurately up to 3-4 decimals with the suggested method. The marked row corresponds to the torus defined for images of Fig. 4, where magnitudes vary in proportion to 1:2:1, as they should.

These results support the view that series of GSTs with properly shaped outer filters $\psi_{kn}$ acts as a Fourier series expansion of tori of orientation fields, even though we cannot expect the orientation field or the underlying original to be as "noise-free". We emphasize that the expansion is done in the space of orientation field, not in the gray value space of the original image.

### B. Finite Expansion of Orientation Fields Radially

Formula (8) reveals that $n$ controls the symmetry of the outer scale filters independently of their thickness (controlled by $\mu$), and the radius $r_k$ of the (magnitude) peak, Fig. 10 *Left*. By derivation, it is straight forward to show that the radii $r_k$ are determined as

$$r_k = \sqrt{\mu}\sigma_k \quad (14)$$

Accordingly, we can change the peak location by changing $\mu$ and $\sigma_k$. Here, we will place the $r_k$ concentrically around the keypoint and in an equidistant manner on the $\log r$ scale so that they will progress geometrically with a constant (design) factor $\alpha = r_{k+1}/r_k$.

As detailed in the Appendix, the parameter $\mu$ alone determines all properties related to the thickness of $\psi_{kn}$: the intersection location $\tau$ of successive tori, or the filter attenuation (height at the next torus peak location) $\tau_\epsilon$. These parameters $\tau, \tau_\epsilon$ are indicated in Fig. 10 and can be used in the filter design process directly, see the Appendix.

### C. Symmetry Assessment by Finite Expansion–Feature Vector

In this section we present our feature set which describes the normalized orientation field $I_{20}^L$ (or the frequency field $I_{20}^F$), around a keypoint by projecting the fields on the set of complex filters presented above. The projection coefficients $c_{kn}$ have precise geometrical meanings via their interpretation as the orientation components ($I_{20}^{(n)}$) of GSTs detecting presence of curves described by harmonic function pairs.

In particular, we suggest $\tilde{h}_k$ defined as orientation field premultiplied with a torus

$$\tilde{h}_k = I_{20}^L \cdot |\psi_{kn}| \quad (15)$$

as the data to be described by GST projections. The projection coefficients are then obtained as

$$c_{kn} = I_{20}^{(k,n)} = <\psi_{kn}, |\psi_{kn}| \cdot I_{20}^L> = <|\psi_{kn}|^2 e^{-in\varphi}, I_{20}^L> \quad (16)$$

The superscript of the $I_{20}$ component of the GST now contains $k$, the torus identity, in addition to $n$.

We introduce the Symmetry Assessing Finite Expansion, *SAFE*, descriptor for keypoints as the ratio comprised of $K \times N$ elements, denoted by

$$SAFE_{kn} = \frac{c_{kn}}{e_k} \in \mathbb{C} \text{ with } k \in \mathbb{N}^+, n \in \mathbb{Z} \quad (17)$$

The descriptor elements represent thus the projection of the orientation field in the $k$'th torus, onto the $n$'th harmonic basis function, as a "fraction" of the contrast of the torus. The quote is motivated by that $c_{kn}$ is complex and the term actually refers to the (real and non-negative) fraction $|c_{kn}|/e_k$. Because the basis set $\psi_{kn}$ is complete, the space and symmetry dimension can be varied systematically to adapt the description power of its finite subsets to the application at hand. The argument $\angle SAFE_{kn} = \angle c_{kn}$ is real and it therefore (continuously) points out which member of the symmetry family (pointed at by the integer $n$) stands for the explanation.

### D. Rotation invariance by steering the features–not the image

The continuous descriptor suggested above allows for rotation invariance of the feature without rotating neither the original image, nor the orientation field. One feature vector can be rotated towards another[7] directly in the feature space

$$SAFE'_{kn} = e^{i(n+2)\varphi'} SAFE_{kn} \quad (18)$$

if the original image is rotated with the angle $\varphi'$. Projection on the spiral pattern $n = -2$ requires thereby no rotation compensation, which is correct, since neither $c_{k,-2}$ nor $e_k$ change when the keypoint rotates.

As an example, drawn from forensic fingerprint images, the minutiae directions will be available between the keypoints of the reference and the query images. To match both minutiae (keypoints) one can match their SAFE vectors. However, before doing that, one of the SAFE vectors can be rotation steered towards the other (with the difference of their minutiae angles as $\varphi'$). If such directions are not available in other applications, one can use an intrinsic orientation of a key point which can be defined to be one of the angles deduced from a component, $\angle c_{kn}$. The intrinsic angle that corresponds to patterns resembling parabolas is $\angle SAFE_{k,-1}$, and is unique as opposed to delta pattern which has 3-folded ambiguity in orientation. Its direction coincides with the minutia direction in the smallest tori, representing the scale of the ridge ending or bifurcation, i.e. if this direction is not automatically extracted for tenprints by other means, it can be extracted by $\angle c_{k,-1}$ in an appropriate scale, [33].

### E. Built-in quality measurements by tight inequalities

Applying the triangle inequality to (16), and remembering that $\|\psi_{kn}\|^2 = 1$ and $|I_{20}^L| \leq 1$, the inequality

$$|c_{kn}| \leq 1 \quad (19)$$

can be obtained. The inequality is tight because $|c_{kn}| = 1$ iff i) the orientation field is reliable, i.e. $|I_{20}^L| = 1$ *everywhere* in the entire torus area, and ii) $\psi_{kn}$ can explain the orientation field $I_{20}^{(kn)}$ without error, $n\varphi = \angle I_{20}^L$. Thus, by way of example, if $|c_{kn}| = 1$, all orientation field data in the torus (support) are reliable, and the $n$'th symmetry basis $e^{-in\varphi}$ can explain them fully. However, if $|c_{kn}| = 0.5$, we can not know if this is due to lack of reliable data in half of the torus (support) or if it is because $n$'th symmetry basis cannot fully explain the

---

[7]Alternatively both are rotated towards the same reference angle.

orientation field within the torus. Accordingly, $|c_{kn}|$ stands for the amount of reliable orientation field within the torus which can be explained by $n$'th symmetry basis, 1 being *full* explanation of the *entire* torus. To disambiguate the interpretation, we use $e_k$ for depicting the amount of reliable orientation within torus, 1 being the *entire* torus.

Using the inequalities, ($e_k \leq 1$), ($|I_{20}^L| \leq 1$) and (19) yields

$$|SAFE_{kn}| \leq 1 \quad (20)$$

The inequality is tight, thus $|SAFE_{kn}|$ represents the amount of reliable orientation field within a *subset* of the torus $k$, explained by the $n$'th symmetry basis, 1 being full explanation.

### F. Matching descriptors

To match two keypoints, the reference keypoint with a test keypoint, their respective complex descriptor arrays, $SAFE^r$ and $SAFE^t$ can be matched, assuming that rotation compensation was made if necessary, Sec. III-D. Using an ordinary scalar product for the complex Euclidean space,

$$<SAFE^r, SAFE^t> = \sum_{kn} SAFE_{kn}^{r*} \cdot SAFE_{kn}^t \quad (21)$$

a complex matching score $MS$ can be defined by

$$MS = \frac{<SAFE^r, SAFE^t>}{<|SAFE^r|, |SAFE^t|>} \Rightarrow |MS| \leq 1 \quad (22)$$

The inequality concerning its magnitude holds and is tight due to the triangle inequality. The equality $|MS| = 1$ holds iff $SAFE^r = z_0 SAFE^t$ with $z_0$ being a non-trivial complex constant, i.e. $\angle SAFE_{kn}^r = \angle z_0 + \angle SAFE_{kn}^t$ and $|SAFE_{kn}^r| = |z_0||SAFE_{kn}^t|$, for all components $k, n$. In order for two descriptors to match the angle between them should vanish, $\angle z_0$. Therefore, we must require that $|MS|$ is high *and* $\angle MS = 0$. This is achieved by using the real part of $MS$ score to embody both magnitude and angle into the final matching score

$$\Re(MS) = |MS|\cos(\angle MS) \in [-1,1] \quad (23)$$

where 1 represents full match, -1 full miss-match, 0, uncertainty. Low or zero certainty happens when the certainties in one of the respective descriptor components (their magnitudes) are zero, because of low quality data in torus or if the orientation data of reliable sectors of torus cannot be explained by the respective symmetric pattern, $n$. Full miss-match, $-1$, occurs when reliable sectors $|MS| = 1$ of all components point at member patterns that are locally orthogonal.

## IV. EXPERIMENTS

First, we report on the specifics of the filters. Then we present our results based on two applications illustrating the performance of the suggested image descriptors.

## A. Filters

The filters we used for extracting SAFE features were designed empirically but guided by the application. First, we determined 10 tori peak-locations as a geometric progression $r_k = r_0 \alpha^k$ with $k = 1 \cdots 9$. This determines the peak locations of the 9 tori without ambiguity (including $\alpha = 1.54$, see Appendix. For fingerprint application the range is fixed to be from $r_0 = 2$ to $r_9 = 97$ while for periocular application it is tested in proportion to sizes of pupil and sclera.

Second, we fixed the attenuation constant as $\tau_\epsilon = 0.01$, Fig. 10, to assure that all (but one) filters were in practice vanishing by the next tap of torus peaks. Fixing attenuation, rather than fixing the intersection height of filters with neighbors, was more practical from implementation point of view. The torus parameters were then available (as $\mu = 20$, and $\sigma_k = r_k/\sqrt{\mu}$), Appendix Lemma 1. The expression of $\mu$ is independent of $k$ which is a consequence of the suggested construction of filter series $\psi_{kn}$ with negligible overlap between them. Finally, we have determined the 9 pattern families deduced from a systematic change of the symmetry index $n \in [-4, 4]$.

## B. Application to Forensic fingerprints

Justice courts do not accept automatic identification of fingerprints, but rely on forensic examiners. Such an expert extracts keypoints such as minutiae, cores or deltas manually from a fingermark and verifies them against those of tenprints suggested by an automatic matcher. Currently, only keypoint constellation (keypoints locations, directions, and types when available) are used in the automatic matching subserving the experts.

We report results of matching via statistics of False Acceptance (FA) rate, False Rejection (FR) rate, Equal Error Rate, and Cumulative Matching Curve (CMC). The first three represent a verification scenario, whereas the last represents an identification scenario. There is a theoretical connection of FA+FR rates with CMC curves [34], e.g. the latter can be obtained from the first, but not vice-versa. Also, CMC statistics are percentages of the background database whereas (derivatives of) FA and FR are impostor and client score distributions. This makes the CMC statistics to scale with the size of the background dataset whereas FA and FR tend to remain less sensitive, since distributions are normalized with the size of the data sets.

We have tested SAFE feature vectors for keypoints to quantify their description power independent of keypoint constellation using the SD27 database of NIST, USA, [35]. It is a data set where keypoints have been annotated by fingermark experts in USA on 258 genuine (matching) tenprint-fingermark pairs. Although the details of the annotation (concerning the matching keypoints) are available by displaying them and visually inspecting them on images of tenprint-fingermark pairs, the (same) correspondence is not available (to computers) at the keypoint level in the original dataset of NIST.

We remedied this by isolating the corresponding keypoints and attributing unique labels to them, (*keypoint identities*) in a recent study, [36]. The thus established (ASCII) correspondence of 5,449 minutiae pairs and 262 cores (*match set*) distributed over 201 (out of 258) tenprint-fingermark pairs, have been used in the present study as groundtruth. The number of cores is different than the fingerprint pairs because some fingerprint pairs had several cores, whereas some had none in common.

Despite that SD27 has "only" 258 fingerprint pairs, this is a large and important database because i) it offers groundtruth to thousands of keypoint identities ii) it is the visual characteristics of a keypoint which represent the source of identity establishment by fingermarks experts, and iii) it is time-consuming to annotate tenprint-fingermark pairs demanding considerable expert resources.

Our experimental results indicate that SAFE features are stable when computed for large tori, where more vectors "vote" and therefore the resulting features are less noisy. We have therefore used the 3 outermost filters with magnitude peaks at $r_i = [27, 41, 63]$ pixels resulting in 3×9 elements in the descriptor. Applying rotation compensation with angle of keypoints obtained automatically (by SAFE features with $n = -1$) or by the manually marked angles (by expert) have demonstrated similar recognition performance in our experiments. This indicates that SAFE can extract the intrinsic directions of the keypoints reliably and suggest it to the expert for verification in fingermarks, if the expert marks their locations. Automatic location of key points can be done by means of GST too, [37].

We have used cores as keypoints in the first experiment, [38]. The verification and identification performance were thus for individual core identities, and are given in Fig. 6 *Black*. The FA and FR rates are summarized by the EER of 25%. The latter means that 75% of the totality (of the possible) 33,092 imposting cores were correctly found to be so, based only on the image information around them. Using the same decision threshold, 75% of the total 262 cores in the fingermarks were correctly matched. In previous studies only [36] reports on verification performance which can be summarized as 36% EER, using the Bozorth3 [39] based on *only minutiae* of SD27, Fig. 6. Since orientation in core neighborhood and minutiae constellations are highly complementary, if not strictly independent, these figures can be seen as a support for that keypoint orientations have significant potentials to improve minutiae constellations, and vice-versa.

We have also implemented an identification experiment on cores using SAFE features summarized by the CMC graph, Fig. 6 *Right*. The CMC *Black* displays an identification range of 16% to 58% for the rank range of 1-20 and uses all fingermark cores against all tenprint cores, implying 33,092 impostors, upon trying to pull-out each of the 262 (client) cores based on their orientation maps only. The study of [36] reported the identification range of 55-78% by using k-plet matcher [40] for ranks 1-20 based on the minutiae set provided by the *SD27, ideal set*, Fig.6. This set is composed of machine generated minutiae for the tenprints and minutiae of fingermarks provided by human experts (without seeing tenprints, therefore "ideal"). The CMC graph is based on one fingermark against all tenprints in SD27, i.e. it contains 258 client (authentic) verifications and 66,048 (256×258, since one tenprint corresponds to two fingermarks) impostor veri-fications. Similarly, the study [17] reports the corresponding identification rate interval of 63-83% using the same protocol. However, they used a different matcher (the greedy minutiae matcher suggested in the paper) and the minutiae of tenprints were not the same. Their tenprint minutiae were extracted by a commercial software (undisclosed algorithm) for the tenprints, and a subset of the minutiae provided by the human experts of SD27 for the fingermarks were deleted.

Support for a similar conclusion, has been provided also by [41], but using different identification protocols than those of our experiments, involving a commercial software. Nonetheless, the study reports 35% and 50% rank-1 identification, when using minutiae constellation alone and when additionally using core points, respectively. They have not reported rank-20 identification for the same experiment. By including additional features such as cores, deltas, quality maps and orientation, significant gains in recognition performance were found by another study as well, [42], albeit in the context of tenprint-tenprint matches.

We extended our experiments in two ways, [43]. First, we have merged two SAFE feature sets yielding 6×9 elements, one set describing the orientation field (3×9 as above) around a keypoint, and another (also 3×9) describing the ridge frequency (density) field around the same point, Sec. II. Second, we have chosen the keypoints as minutiae (instead of cores), to evaluate if SAFE descriptors could be useful even if cores were not available. As in our previous experiments, we used one or two keypoints (i.e. minutiae instead of cores) per tenprint-fingermark pair in the experiments, resulting in 320 client and 50,978 impostor comparisons in both verification and identification scenarios. It was possible to use SAFE features to describe ridge frequency fields because they can be encoded by complex fields, similar to orientation fields, [28]. The minutiae were chosen such that they were having high orientation variation in their neighborhoods. Distances of the chosen minutiae to cores, if these were present, varied between 2-50 pixels. Additionally the distances of (chosen, expert marked) minutiae to cores did not always agree well between tenprints and fingermarks, varying between 0-250 pixels, mainly due to non-linear distortion in fingermarks. Nonetheless, we have included at least one minutia from all 258 pairs of fingerprint images in the experiment, (resulting in 320 pairs), that is even if a fingermark did not include cores nor had significant orientation variation otherwise.

These experiments showed an improvement in the FA, FR performance summarized by a lower EER, 19% (down from 24%), Fig. 6 *Left*. The identification experiments showed an improvement with the rank-20 correct identification rate of 74% (up from 58%), Fig. 6 *Right*. The outcome indicates that i) SAFE features are not critically dependent on finding cores in fingermarks, and ii) they can be computed for, and merged with other vector fields than genuine orientation fields. It is important to highlight that CMC curves for core and minutia experiments have different data size, therefore we neither can compare identification percentage between each other nor plot curves together on the same coordinate system. All percentages are given to show the tendency, nevertheless, we cannot claim performance improvement as compared to

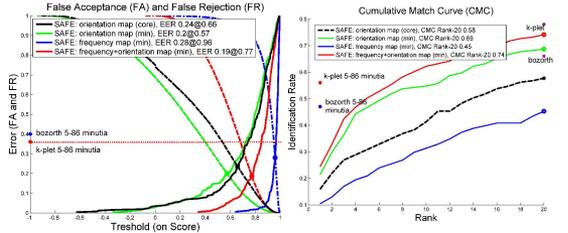

Fig. 6. Performance with Equal Error Rate (EER) and Cumulative matching curve (CMC) on SD27 forensic fingerprint database. *Black*. Matching core points of the fingerprint by orientation based SAFE descriptor. We display CMC of the core experiment together with minutiae ones for brevity. *Red*. Matching minutiae by orientation (*Green*) and frequency (*Blue*) based SAFE descriptors

other algorithms because they have reported only CMC curves.

The study of [18] is conceptually relevant and has results using SD27, but they are more difficult to relate to our results. This is because they have not presented pure minutiae constellation performance. Their CMC reporting is based on minutia plus image information extracted by commercial software (undisclosed). Neither they have reported these without and with onset cores, or image neighborhood information.

*C. Application to iris images for periocular recognition*

In this section, the SAFE feature extractor is tested for periocular recognition on high quality close-up iris images from the BioSec database [44]. We select 1,200 iris images originated from 75 individuals and acquired in 2 sessions (4 images of each eye per person, per session). Images are acquired with a LG IrisAccess EOU3000 close-up infrared iris camera with resolution of 480×640 pixels. BioSec database has been annotated manually [45] such that positions of the center of the pupil/sclera circles and their radius are known. Iris images possess different sources of noise as compared to fingerprints, e.g. eyelashes and eyelids, as well as variations in lighting or view angle [46].

SAFE features are extracted on a grid of points in the periocular area. The grid has rectangular geometry, with sampling points distributed uniformly, and located in the iris center as pictured in Fig. 7 *Left*. This setup is inspired by our previous works on periocular recognition [47]. Matching between two images is done by computing the matching score $\Re(MS)$, eq. (23) between corresponding points of the sampling grid. All matching scores are then averaged, resulting in a single matching score between two given images. Due to the nature of iris close-up acquisition, there is no significant rotation variation between different captures. As a result, we have observed that rotation compensation has no significant improvement with such grid applied to close-up iris images [47], therefore no rotation compensation is used in this case. Each eye of the database is considered as different user, thus having 150 different users. Genuine matches for each user are obtained by comparing all images of the $1^{st}$ session to all images of the $2^{nd}$ session. Impostor matches are obtained by comparing the $2^{nd}$ image of the $1^{st}$ session of a user to the $2^{nd}$ image of the $2^{nd}$ session of the remaining users. This leads to 150×4×4=2,400 genuine and 150×149=22,359 impostor comparisons.

We have compared SAFE features with the Gabor-based periocular system proposed in [47], which uses the same sampling grid as in Fig. 7. In this system, the local power spectrum of the image is sampled at each point of the grid by a set of Gabor filters organized in 5 frequency channels and 6 equally spaced orientation channels, thus resulting in $5 \times 6 = 30$ filter responses per sampling point. Gabor filter wavelengths span from 16 to 60 pixels. This covers approximately the range of pupil radius, as given by the ground-truth [45], see Fig. 7 *Right*. The Gabor responses from all points of the grid are grouped into a single complex vector, which ... identity model. Matching between two given imag... via $\chi^2$ distance of the magnitude of complex va... to matching with magnitude vectors, they are nor... a probability distribution (PDF) by dividing each ... the vector by the sum of all vector elements. So... experiments are also done between different mat... fused distance is computed as the mean value of th... due to the individual matchers, which are first nor... be similarity scores in the $[0,1]$ range using tanh... as $s' = \frac{1}{2}\left\{\tanh\left(0.01\left(\frac{s-\mu_s}{\sigma_s}\right)\right) + 1\right\}$. Here, $s$ is t... distance score and $s'$ is the normalized similarity score, $\mu_s$ and $\sigma_s$ are respectively the mean and standard deviation of the genuine score distribution [48].

Performance of SAFE features ('PP') is given in Fig. 8 *Left*. We also provide results (*Right*) of the Gabor-based periocular ('PG') and the fusion of both matchers ('PP+PG'). The corresponding EERs are given in Table I. The size of the torus has been varied according to average size of the iris, as given by the ground-truth [45], see Fig. 7 *Right*. The smallest filter radius is set proportional to 30 (average pupil radius), and the biggest filter radius is set proportional to 100 (slightly smaller than the average sclera radius). This leads to the combinations '15-100' and '30-200', or a 'big size torus'. We have also tested a 'small size torus' by setting the largest filter radius proportional to 30, with the smallest filter radius reduced accordingly. This leads to the combinations '5-30', '5-60', and '10-60'.

Results of Fig. 8 *Left* show some differences between big and small toruses, but they are not very significant, with EER varying from 12.8% to 14%. These are competitive verification rates in comparison with existing periocular approaches [20], [13], [21], which are between 7% and 22% (depending on the features used). The tendency observed with SAFE features is that performance is slightly better with a smaller torus, with the best configuration corresponding to '5-60'. Only when the top-end of the range of radii is made large in comparison with the iris size (i.e. '30-200'), the performance shows an appreciable worsening. When compared with the Gabor periocular system ('PG'), SAFE features perform worse. However, the fusion of the two systems ('PP+PG') shows an improvement of up to nearly 14%. By being (sampled local) Fourier Transform magnitudes, Gabor magnitude features are invariant to small translations [49], whereas SAFE features are translation sensitive by design. This different behavior can explain the improvement, via the complementarity between these two features observed in our fusion experiments (recall that the two features are extracted from exactly the same points of the image).

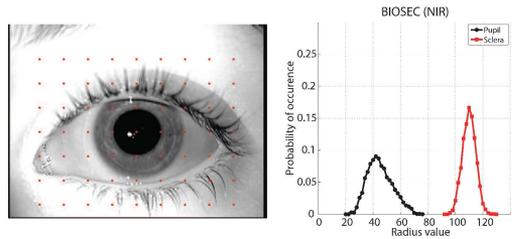

Fig. 7. *Left:* Sampling grid applied to periocular area of iris images of the BioSec database [44]. *Right:* histograms of pupil and sclera radius of the BioSec database, as given by the groundtruth [45].

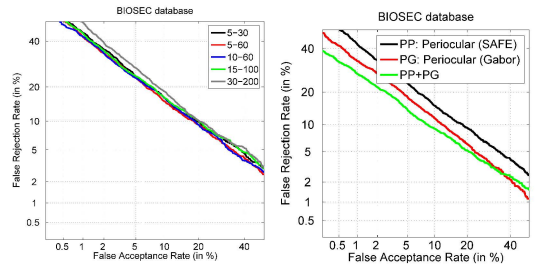

Fig. 8. *Left:* DET curve demonstrating performance of SAFE features on the BioSec database for filters of different sizes. *Right:* DET curve as comparison to the other periocular matcher based on Gabor filters, as well as of fusion experiments.

## V. DISCUSSION

### A. Automatic features and forensic experts

We suggested a model driven approach for feature extraction which incorporates the quality of the extracted information at the output level of the features. Even at the input level, the dense orientation image, the model admits use of quality measures. In our study, we have used automatic extraction of orientation, even for fingermarks, [28] and reported the various performances accordingly. In a real scenario, the forensic expert may examine the automatically suggested dense orientation image as a color image, overlay/display it on the original, and issue a few mouse clicks to correct the erroneously estimated orientations, or reduce their quality (down to possibly zero) if they are unreliable. However, this is not done here for the benefit of repeatability, and comparisons, but it is an intended way of using the suggested features.

The SD27 database contains forensic fingerprints representing genuine orientation blended with background orientation. It means several orientations may occur jointly at certain locations, e.g. a fingermark on a banknote full of graphics drawings. An automatic software may impose a continuous model to make an intelligent guess to capture only the fingerprint orientations, [50], attempting to reject banknote drawings. This is left as future work, to focus the study, since a fingermark expert can correct the orientations too, as explained above.

| filter configuration | BioSec database (NIR) Periocular | | |
|---|---|---|---|
| | Proposed (PP) | Gabor (PG) | Fusion PP+PG |
| 5-30 | 13.07 | | 9.35 (-13.20%) |
| 5-60 | **12.81** | | **9.28 (-13.87%)** |
| 10-60 | 13.12 | 10.77 | 9.44 (-12.33%) |
| 15-100 | 13.47 | | 9.82 (-8.84%) |
| 30-200 | 13.96 | | 9.89 (-8.21%) |

TABLE I
VERIFICATION RESULTS IN TERMS OF EER. THE BEST CASE OF EACH COLUMN IS MARKED IN BOLD. FUSION RESULTS: THE RELATIVE EER VARIATION WITH RESPECT TO THE BEST INDIVIDUAL SYSTEM IS GIVEN IN BRACKETS (ONLY WHEN THERE IS PERFORMANCE IMPROVEMENT).

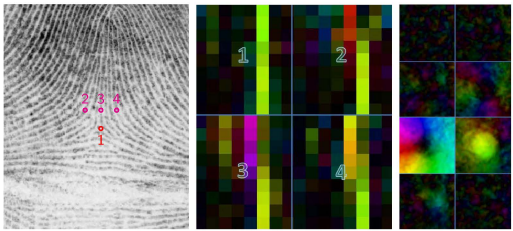

Fig. 9. *Left:* Extraction area with highlighted key points. *Middle:* Feature extracted from center of delta and neighbouring points highlighted on the left. *Horizontal:* The $SAFE_{kn}$ features (as 9x8 matrix) extracted from the marked points (1...4) and represented as complex pixels. *Right* Mosaic of dense maps centered around Point 1 and representing $SAFE_{kn}$ for $n = -4 \cdots 3$ (in reading order) extracted in the (same) torus $k = 7$. All color pixels are complex with angle being hue (estimated parameter) and magnitude being brightness (quality).

We think that the reported results are significant in that without using minutiae constellation information, but by using only (automatically obtained) orientation information around key points (mostly one per fingermark) we were able to obtain a performance similar to constellation of minutiae. Evidently the problem of the interfering background orientation is not attempted to be resolved. However, we think that the features contribute to this in our results, because as said, an expert in the real scenario will be able to assess/correct the orientations around the few key points she/he deems important, before the composite of automatic and manual orientations are encoded into our descriptors.

*B. Object nature of neighborhoods*

The suggested $SAFE_{kn}$ features are complex valued and their magnitudes represent the share of a highly symmetric family in explaining the orientations of an image neighborhood. Examining the iso-curves of a symmetric family $n$, it can be shown that, Fig. 3, except for the symmetric family of $n = 0$, they are singular only at the origin, i.e. their orientations are undefined, at the reference point itself. It means that the filters corresponding to $SAFE_{kn}$ with $n \neq 0$, are characterized by that they seek for evidence of presence of iso-curves which admit the point where the filter is placed, as their sole singularity point. Regardless of $n$, moving the filter to a nearby point will provoke a significantly diminished magnitude of the features $|SAFE_{kn}|$ since there is only one such singularity in the neighborhood. The fact that all symmetric families ($n \neq 0$) agree on where this singularity point is (the origin) makes $SAFE_{kn}$ with $n \neq 0$ sensitive to the "object nature" of an image neighborhood of a neighborhood. Such neighborhoods have an intrinsic origin making their locations precise and unique. Even their global orientation is unique, but up to $n$ folded ambiguity including the patterns with $n = 0$ (ordinary lines sharing an orientation) but excluding those with $n = -2$ (log-spirals are invariants of rotation and zooming). Thus, the features, $n \neq 0$, should be particularly good in giving an intrinsic identity to a point, supposing that it is not a texture point, via finite expansion, and locate it by the virtue of the singularity of the underlying iso-curves.

The quantity $|SAFE_{k0}|$ is large no matter where the corresponding filter is placed inside line patterns sharing a common direction, texture points, Fig. 9. There is no unique point, intrinsic to the pattern, where the magnitude is large and small elsewhere. It is large everywhere, by the nature of textures. Thus $SAFE_{k0}$, the ordinary structure tensor delivering orientation estimates, captures the "texture nature" of the neighborhood since the it is translation invariant. However, it is not as good as describing the "object nature" since there is only one component, $SAFE_{n0}$. To use symmetry derivatives of Gaussians as texture descriptors is possible, [51], but it is outside the scope of the present paper since we are interested in what makes points unique, not what makes them anonymous.

We illustrate how feature extraction differs for "object nature" from the "texture nature" by a fingerprint, Fig. 10. We extract $SAFE_{kn}$ at the center of the delta, 1, and at points 2, 3 and 4. The green column in image 1 has highest quality, i.e. $SAFE_{kn}$ with $n = 1$ (delta type iso-curves) is the largest in all 9 tori around the point. By contrast, for other points (the images 2, 3, and 4), it is the column with $n = 0$ which is the brightest, at least within small tori (k=1,2,3), i.e. these points have texture properties ($n = 0$). If we would move points 2, 3 and 4 around point 1 we should have the same quality (no reduction of brightness), but only change of the hue. This is observable in the fifth image (reading order) of the image of the mosaic on *Right*. It displays the dense map of $n = 0$ (for the 7'th torus) at each image point (zoomed on Point 1). The centre is dark i.e. point 1 is not a good texture point. If we would translate point 1, $SAFE_{kn}$ with $n = 1$ (delta type iso-curves) should signal a lower quality (become darker) since the point has object property (of type $n = 1$) rather than a texture. This is observed in the sixth zoomed image (where $n = 1$) on *Right*, as a bright yellowish spot.

VI. CONCLUSION

We have presented a model-based feature vector to represent the neighborhoods of keypoints via their orientation fields, for recognition purposes. Being model-based allows to have built-in quality measurements for individual descriptors.

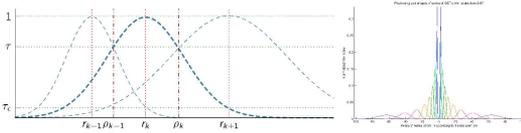

Fig. 10. *Left:* Construction of geometrically positioned series of filters *Right:* Radial functions of filters. Due to normalization larger rings will have the same area (importance) as small ones.

Experimentally we could obtain support for its promising recognition power in isolation from other features, using publicly available data sets in fingerprint forensics as well as periocular biometrics, both subserving identity recognition.

The results are encouraging beyond ROC/CMC curves since the features are generic i.e. not application specific. The basis functions of the model can be adapted to novel applications thanks to the design friendly tori, enabled by the lemmas. The descriptors (except $n=0$) encode location sensitive object properties, offering complementarity to translation invariant texture properties, which prevail current generic feature vectors. The feature space can also be made rotation invariant or rotation compensated easily by complex multiplication, without rotating the input data.

Descriptors having symmetry index $n=-2$ are rotation invariant as they are (no rotation compensation). The GST theory suggests that this feature is scale invariant too. However we have not exploited here to compensate the features against severe scale changes due to our applications, although this is possible by further processing of the descriptors.

## Appendix A
## Filter support properties

The filter function $\psi_{kn}$, (8), is a polar separable 2D function. Defining the radial part as $t$,

$$t(r,\mu,\sigma^2) = r^\mu e^{-\frac{r^2}{2\sigma^2}} \qquad (24)$$

its maximum $C$ is reached at $r=\sqrt{\mu}\sigma$

$$C = \sup_r t(r,\mu,\sigma^2) = t(\sqrt{\mu}\sigma,\mu,\sigma) = (\sqrt{\mu}\sigma)^\mu e^{-\frac{\mu}{2}}. \qquad (25)$$

For our experiments we used a set of filters with peak locations $r_k$, growing in geometric progression with the constant factor $\alpha$ so that $r_k = r_0 \alpha^k$ with $k \in 0 \cdots K-1$, Fig. 10 *Right*. The purpose is two folded: i) to achieve spatial completeness by increasing K in a bounded area around a keypoint, and ii) to preserve more of the orientation variations close to the keypoint via thinner tori, since a torus with a small $k$ is both closer to the keypoint and thinner than one far away. The location of a filter peak $r_k$ is thus controlable via $\sigma = \sigma_k$

$$r_k = \sqrt{\mu_k}\sigma_k \Longrightarrow \sigma_k = \frac{r_0 \alpha^k}{\sqrt{\mu_k}} \qquad (26)$$

if $\mu_k$ is known, in addition to the design parameters $K$, $r_0$ and $\alpha$.

The peak locations are possible to determine only if $\mu_k$ are known. What property of the filter does a choice of $\mu_k$ steer then? The answer to this is given by the following lemma and amounts to that $\mu_k$ determines the degree of overlap between subsequent filters and can be controlled by fixing $\tau_\epsilon$ and $\alpha$.

**Lemma 1.** *The values of a sequence of peak-normalized filter magnitudes $t(r,\mu_k,\sigma_k^2)/C_k$ depreciate to the same value (height) $\tau_\epsilon$ at the location of the next filter peak, $r_{k+1}$, independent of $k$, provided that the peak locations are in geometric progression with a constant factor $\alpha = \frac{r_{k+1}}{r_k}$. This (common) height $\tau_\epsilon$ determines the real constant $\mu > 0$ and vice-versa via*

$$\mu = \frac{\log \tau_\epsilon}{\log \alpha - \frac{\alpha^2-1}{2}}, \qquad (27)$$

*so that $\mu_k = \mu$, and $\mu/\log\tau_\epsilon$ is constant for all $k$.*

*Proof.* Calling a peak normalized filter magnitude $\tilde{t}(r,\mu_k,\sigma_k^2) = t(r,\mu_k,\sigma_k^2)/C_k$ and fixing its peak location at $r_k = \sqrt{\mu_k}\sigma_k$ determines $\sigma = \sigma_k$, (25) and (26). This filter's magnitude $\tau_\epsilon$ at $r_{k+1}$ is then given by

$$\tau_\epsilon = \tilde{t}(r,\mu_k,\frac{r_k^2}{\mu_k}) = \frac{r^{\mu_k} e^{-\frac{\mu_k r^2}{2r_k^2}}}{r_k^{\mu_k} e^{-\frac{\mu_k}{2}}} = (\frac{r}{r_k})^{\mu_k} e^{\frac{\mu_k}{2}} e^{-\frac{\mu_k}{2}(\frac{r}{r_k})^2}$$

However, at $r=r_{k+1}$ the quotient $\frac{r}{r_k} = \alpha$ is given as constant. The value of $\tilde{t}$ is then

$$\tau_\epsilon = \alpha^{\mu_k} e^{-\frac{\mu_k}{2}(\alpha^2-1)} = (\alpha e^{-\frac{\alpha^2-1}{2}})^{\mu_k} \qquad (28)$$

Inverting the equation w.r.t. $\mu_k$, whereby $\mu_k$ becomes independent of $k$ since $\tau_\epsilon$ is constant, achieves the remainder of the proof. □

There exist situations in which steering the amount of overlap is more convenient via $\tau$ than $\tau_\epsilon$, Fig. 10 *Left*, for example when they are to be used in conjunction with subsampling in pyramid processing. The question is if $\tau$ too can steer $\mu_k$ freely and yet the latter is independent of $k$. The answer is yes as precised by the next lemma which furthermore concludes that $\mu_k$ even determines the amount of overlap between subsequent filters.

**Lemma 2.** *The values of a sequence of peak-normalized filter magnitudes $t(r,\mu_k,\sigma_k^2)/C_k$ with peak locations $r_k$, depreciate to the same height at the intersection with the next filter in the sequence, provided that the peak locations are in geometric progression with a constant factor $\alpha = \frac{r_{k+1}}{r_k}$. This height $\tau$ determines the real constant $\mu > 0$ and vice-versa via*

$$\mu = \frac{\log \tau^2}{\log[\log(\beta\beta^{-1})]+1}, \text{ where } \beta = \alpha^{\frac{2}{\alpha^2-1}} \qquad (29)$$

*so that $\mu_k = \mu$, and $\mu/\log\tau^2$ is constant for all $k$.*

*Proof.* Assuming $\mu_k = \mu$ i.e. a constant independent of $k$, the equation $\tilde{t}(\rho_k,\mu,\frac{r_k^2}{\mu}) = \tilde{t}(\rho_k,\mu,\frac{r_{k+1}^2}{\mu})$ is established to obtain the intersection location $\rho_k$

$$\frac{C_k}{C_{k+1}} = \exp[-\mu \frac{\rho_k^2}{2r_0^2}(\frac{1}{\alpha^{2k}} - \frac{1}{\alpha^{2k+2}})] \qquad (30)$$

from which $\rho_k$ is solved as

$$\rho_k = r_0 \sqrt{\frac{\log \alpha^2}{\alpha^2-1}} \alpha^{k+1}. \qquad (31)$$

Accordingly, the intersection locations are in geometric progression with the same factor $\alpha$ as the peak locations.

Next, $\rho_k$ is substituted into $\tilde{t}$ to obtain the filter magnitudes at the intersections as

$$\tau = \tilde{t}(\rho_k, \mu, \frac{r_k^2}{\mu}) = r_0^\mu (\frac{\log \alpha^2}{\alpha^2 - 1})^{\frac{\mu}{2}} \alpha^{\mu(k+1)} e^{-\frac{\mu}{2} \frac{\rho_k^2}{r_k^2}}$$

which simplifies to

$$\tau = (\frac{\alpha^{\frac{2}{\alpha^2-1}} \log \alpha^2}{\alpha^2 - 1})^{\frac{\mu}{2}} e^{\frac{\mu}{2}} \qquad (32)$$

By inverting the expression w.r.t. $\mu$, eq. (29), wherein $\tau$ and $\alpha$ are constants, confirming existence of $\mu$ independent of $k$, is obtained. □